\documentclass[runningheads]{llncs}
\usepackage{graphicx}

\usepackage{tikz}
\usepackage{comment}
\usepackage{amsmath,amssymb} 
\usepackage{color}

\usepackage[accsupp]{axessibility}  
\usepackage{authblk}
\makeatletter
\renewcommand\AB@affilsepx{, \protect\Affilfont}
\makeatother
\usepackage{tikz}
\usepackage{comment}
\usepackage{amsmath,amssymb}
\usepackage{color}
\usepackage[accsupp]{axessibility}
\usepackage{cite}
\usepackage{xurl}
\usepackage{epsfig}
\usepackage{diagbox}
\usepackage{array}
\usepackage{booktabs}
\usepackage{nth}
\usepackage{nicefrac}
\usepackage{relsize}
\usepackage{xspace} 
\usepackage{bm}
\usepackage{enumerate}
\usepackage{adjustbox}
\usepackage{multirow}
\usepackage{paralist}

\makeatletter
\newcommand{\thickhline}{%
    \noalign {\ifnum 0=`}\fi \hrule height 1pt
    \futurelet \reserved@a \@xhline
}

\usepackage{cuted}
\usepackage{capt-of}
\newcolumntype{P}[1]{>{\centering\arraybackslash}p{#1}}
\newcommand{\mytilde}{\raise.17ex\hbox{$\scriptstyle\mathtt{\sim}$}}

\usepackage[pagebackref,breaklinks,colorlinks,linkcolor=blue,citecolor=blue,urlcolor=blue]{hyperref}
\usepackage[capitalize]{cleveref}
\crefname{section}{Sec.}{Secs.}
\Crefname{section}{Section}{Sections}
\Crefname{table}{Table}{Tables}
\crefname{table}{Tab.}{Tabs.}

\begin{document}
\pagestyle{headings}
\mainmatter
\def\ECCVSubNumber{9}  

\makeatletter
\renewcommand*{\@fnsymbol}[1]{\ifcase#1\or*\else$\dagger$\fi}
\makeatother

\makeatletter
\newcommand{\printfnsymbol}[1]{%
  \textsuperscript{\@fnsymbol{#1}}%
}
\makeatother

\title{Pixel2ISDF: Implicit Signed Distance Fields based Human Body Model from Multi-view and Multi-pose Images} 


\titlerunning{Pixel2ISDF}
%
\author{Jianchuan Chen\thanks{J. Chen, W. Yi, T. Wang and X. Li - Equal contributions.}\inst{1} \and
Wentao Yi\printfnsymbol{1}\inst{1} \and
Tiantian Wang\printfnsymbol{1}\inst{2} \and Xing Li\printfnsymbol{1}\inst{3} \and Liqian Ma\inst{4} \and Yangyu Fan\inst{3} \and Huchuan Lu\thanks{Corresponding author.}\inst{1}}
\authorrunning{J. Chen et al.}
%
\institute{Dalian University of Technology, China \\
\and University of California, Merced \\
\and Northwestern Polytechnical University, China\\
\and ZMO AI Inc.\\
\email{\{Janaldo, raywit\}@mail.dlut.edu.cn}, \email{lhchuan@dlut.edu.cn}\\
\email{twang61@ucmerced.edu},
\email{fan\_yangyu@nwpu.edu.cn}\\
\email{lixing36@foxmail.com}, \email{liqianma.scholar@outlook.com}
}
\maketitle

\renewcommand{\thefootnote}{\fnsymbol{footnote}} 

\begin{abstract}

In this report, we focus on reconstructing clothed humans in the canonical space given multiple views and poses of a human as the input.
To achieve this, we utilize the geometric prior of the SMPLX model in the canonical space to learn the implicit representation for geometry reconstruction.
Based on the observation that the topology between the posed mesh and the mesh in the canonical space are consistent, we propose to learn latent codes on the posed mesh by leveraging multiple input images and then assign the latent codes to the mesh in the canonical space.
Specifically, we first leverage normal and geometry networks to extract the feature vector for each vertex on the SMPLX mesh. Normal maps are adopted for better generalization to unseen images compared to 2D images. Then, features for each vertex on the posed mesh from multiple images are integrated by MLPs.
The integrated features acting as the latent code are anchored to the SMPLX mesh in the canonical space. Finally, latent code for each 3D point is extracted and utilized to calculate the SDF. Our work for reconstructing the human shape on canonical pose achieves 3rd performance on WCPA MVP-Human Body Challenge.

\end{abstract}

\section{Introduction}

High-fidelity human digitization has attracted a lot of interest for its application in VR/AR, image and video editing, telepresence, virtual try-on, etc. 
In this work, we target at reconstructing the high-quality 3D clothed humans in the canonical space given multiple views and multiple poses of a human performer as the input. 

Our network utilizes multiple images as the input and learns the implicit representation for the given points in the 3D space. Inspired by the advantages of implicit representation such as arbitrary topology and continuous representation, we adopt this representation to reconstruct high-fidelity clothed 3D humans. To learn geometry in the canonical space, we utilize the SMPLX mesh~\cite{loper2015smpl} in the canonical space as the geometric guidance. Due to the correspondence between the posed mesh and the mesh in the canonical space, we propose to first learn the latent codes in the posed mesh and then assign the latent codes to the canonical mesh based on the correspondence. By utilizing the posed mesh, image information in the 2D space can be included by projecting the posed mesh to the image space.

Given the multi-view images as input, normal and geometry networks are utilized to extract the features for the vertices on the SMPLX mesh~\cite{loper2015smpl}. We utilize a normal map as the intermediate prediction which helps generate sharp reconstructed geometry~\cite{saito2020pifuhd}.


To integrate the features from different views or poses, we utilize a fusion network to generate a weighted summation of multiple features. Specifically, we first concatenate the features with the means and variances of features from all inputs followed by a Multi-layer Perceptron (MLP)  predicting the weight and transformed features. Then, the weighted features will be integrated into the latent code through another MLP.

The latent code learned by the neural network is anchored to the SMPLX mesh in the canonical space, which serves as the geometry guidance to reconstruct the 3D geometry. Because of the sparsity of the vertices, we utilize a SparseConvNet to generate effective features for any 3D point following~\cite{peng2020neural}. Finally, we use the trilinear interpolation to extract the latent code followed by an MLP to produce the SDF which models the signed distance of a point to the nearest surface.

\section{Related Work}

\noindent \textbf{Implicit Neural Representations.} Recently, the implicit representation encoded by a deep neural network has gained extensive attention, since it can represent continuous fields and can generate the details on the clothes, such as wrinkles and facial expressions. 
Implicit representation has
been applied successfully to shape representation~\cite{mescheder2019occupancy,park2019deepsdf,yifan2021iso}.
Here we utilize the signed distance function (SDF) as the implicit representation, which is a continuous function that outputs the point’s signed distance to the closest surface, whose
sign encodes whether the point is inside (negative) or outside (positive) of the watertight surface.
The underlying surface is implicitly represented by the isosurface of $SDF(\cdot) = 0$.

\noindent \textbf{3D Clothed Human Reconstruction.} Reconstructing humans has been widely studied given the depth maps\cite{chibane2020implicit,wang2020normalgan,yu2018doublefusion}, images~\cite{bogo2016keep,kanazawa2018end,kolotouros2019learning}, or videos~\cite{kanazawa2019learning,kocabas2020vibe} as the input. Utilizing the SMPLX mesh, the existing works show promising results with the RGB image as the input.
NeuralBody~\cite{peng2020neural} adopts the posed SMPLX mesh to construct the latent code volume that aims to extract the implicit pose code. ICON~\cite{xiu2022icon} proposes to refine the SMPLX mesh and normal map iteratively. SelfRecon~\cite{jiang2022selfrecon} utilizes the initial canonical SMPLX body mesh to calculate the canonical implicit SDF. Motivated by these methods, we propose to utilize the SMPLX mesh in the canonical space as a geometric prior to reconstruct the clothed humans.

\section{Methodology}
Given the multi-view and multi-pose RGB images of human and SMPLX parameters as the input, we aim to reconstruct the clothed 3D geometry in the canonical space.
Here the input images are denoted as $\{I_k\}_{k=1}^{K}$, where $k$ denotes the image index and $K$ is the number of images. 
The corresponding SMPLX parameters are denoted as $\{\theta_k,\beta_k, s_k, t_k\}_{k=1}^{K}$, where $\theta_k,\beta_k$ are the pose and shape parameters of SMPLX and $s_k, t_k$ are the camera parameters used for projection.

\begin{figure}[t]
\centering
\includegraphics[width=0.98\linewidth]{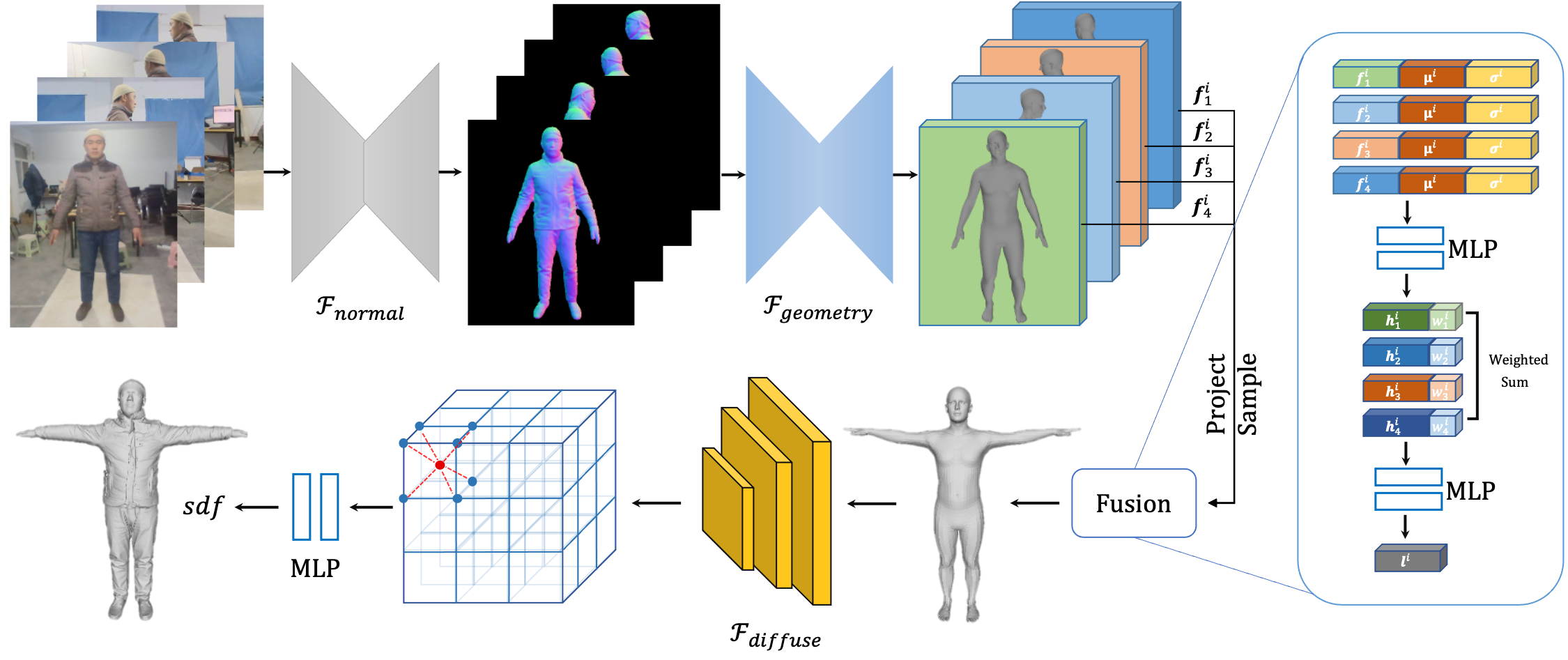}
\vspace{-2mm}
\caption{\textbf{Overview of our clothed human reconstruction pipeline.}}
\label{fig:pipeline}
\end{figure}

\subsection{Multi-view and Multi-pose Image Feature Encoding}
Our feature extraction networks utilizes multi-view and multi-pose images $\{{I_k}\}_{k=1}^{K}$ as input and outputs the geometric feature maps that help to predict the 3D geometry in the canonical space. Specifically, we first adopt the normal network $\mathcal{F}_{normal}$ to extract the normal maps and then utilize the geometry network $\mathcal{F}_{geo}$ to generate geometric feature maps $\{{F_k}\}_{k=1}^{K}$ that will be further utilized to extract the pixel-aligned features for the vertices on the posed mesh.

\begin{equation}
F_{k}=\mathcal{F}_{\text {geometry }}\left(\mathcal{F}_{\text {normal }}\left(I_{k}\right)\right) \quad k=1,2, \cdots, K
\end{equation}

In particular, we adopt the pretrained image-to-image translation network from PIFuHD~\cite{saito2020pifuhd} as our normal network, and use the pretrained ResUNet34~\cite{he2016deep} backbone as our geometry network.



\subsection{Structured Latent Codes}

After obtaining the geometric feature maps $\{{F_k}\}_{k=1}^{K}$, we extract the pixel-aligned features for each vertex $\mathbf{v}_{k}^{i}$ on the posed mesh $M(\theta_k, \beta_k)$. For each vertex $\mathbf{v}_{k}^{i}$, we first project it to the image space by utilizing the weak-perspective projection $\Phi$ according to camera parameters scale $s_k$ and translation $t_k$, then adopt the bilinear interpolation operation to extract the pixel-aligned features $\boldsymbol{f}_{k}^{i}$.

\begin{equation}
\Phi: \hat{\mathbf{x}}=s \Pi(\mathbf{x})+t, \quad \mathbf{x} \in \mathbb{R}^{3}, \hat{\mathbf{x}} \in \mathbb{R}^{2}
\end{equation}

\begin{equation}
\boldsymbol{f}_{k}^{i}=\text {bilinear}\left(F_{k}, \Phi\left(\mathbf{v}_{k}^{i}, s_{k}, t_{k}\right)\right)
\end{equation}
where $\Pi$ is the orthogonal projection, $\mathbf{x}$ is the point in 3D space and $\hat{\mathbf{x}}$ is the projected points in 2D image space.

To integrate the feature of the $i$-th vertex in the canonical space from multiple views/poses, we use a fusion network that takes $\{\boldsymbol{f}_k^i\}_{k=1}^{K}$ as the input and outputs the integrated feature $\boldsymbol{l}^i$, which is illustrated in Figure~\ref{fig:pipeline}. Specifically, the mean $\boldsymbol{\mu}^i$ and variance $\boldsymbol{\sigma}^i$ of features $\{\boldsymbol{f}_k^i\}_{k=1}^{K}$ is calculated and then concatenated with $\boldsymbol{f}_k^i$ to serve as the input of a MLP. 
The MLP predicts the new feature vector and weight $\{
w_k^i\}_{k=1}^{K}$ for each feature, which generates a weighted sum of features from multiple inputs. The weighted feature is finally forwarded to an MLP for feature integration, ${\boldsymbol{l}^i}$ which serves as the structured latent code for the vertex $\mathbf{v}^i$.

Different from the latent code in NeuralBody~\cite{peng2020neural} which is initialized randomly for optimizing specific humans, our latent code is the feature vector learned by a network with the normal map as the input, which can generalize to humans unseen from training ones.

The above-mentioned latent codes are generated based on the posed mesh. The posed mesh and the canonical mesh share the same latent codes because of their topology correspondence which forms the set of latent codes by $\mathcal{Q}=\{
\boldsymbol{l}^1, \boldsymbol{l}^2, \cdots, \boldsymbol{l}^{N_V}\}, \boldsymbol{l}^{i} \in \mathbb{R}^d$. Here $N_V$ represents the number of vertices.

\subsection{Implicit Neural Shape Field} 
The learned latent codes are anchored to a human body model (SMPLX~\cite{loper2015smpl}) in the canonical space. 
SMPLX is parameterized by shape and pose parameters with $N_V = 10, 475$ vertices and $N_J = 54$ joints.
The locations of the latent codes $\mathcal{Q}=\{
\boldsymbol{l}^1, \boldsymbol{l}^2, \cdots, \boldsymbol{l}^{N_V}\}$ are transformed for learning the implicit representation by forwarding
the latent codes into a neural network

To query the latent code at continuous 3D locations, trilinear interpolation is adopted for each point. However, the latent codes are relatively sparse in the 3D space, and directly calculating the latent codes using trilinear interpolation will generate zero vectors for most points.
To overcome this challenge, we use a SparseConvNet~\cite{peng2020neural} to form a latent feature volume $\mathcal{V} \in \mathbb{R}^{H \times H \times H \times d} $ which diffuses the codes defined on the mesh surface to the nearby 3D space.

\begin{equation}
\mathcal{V} = \mathcal{F}_{diffuse}\left(\{\boldsymbol{l}^1, \boldsymbol{l}^2, \ldots, \boldsymbol{l}^{N_V}\} \right)
\end{equation}

Specifically, to obtain the latent code for each 3D point, trilinear interpolation is employed to query the code at continuous 3D locations.

Here the latent code will be forwarded into a neural network $\varphi$ to predict the SDF $\mathcal{F}_{sdf}(\mathbf{x})$ for 3D point $\mathbf{x}$.

\begin{equation}
\mathcal{F}_{s d f}(\mathbf{x})=\varphi(\text { trilinear }(\mathcal{V}, \mathbf{x}))
\end{equation}

\subsection{Loss Function} 
During training, we sample $N_S$ spatial points $\mathcal{X}$ surrounding the ground truth canonical mesh. To train the implicit SDF, we deploy a mixed-sampling strategy: 20\% for uniform sampling on the whole space and 80\% for sampling near the surface. We adopt the mixed-sampling strategy because of the following two reasons. First, sampling uniformly in the 3D space will put more weight on the points outside the mesh during network training, which results in overfitting when sampling around the iso-surface. Second, sampling points far away from the reconstructed surface contribute little to geometry reconstruction, which increases the pressure of network training. 

Overall, we enforce 3D geometric loss ${L} _{sdf}$ and normal constraint loss ${L} _{normal} $.

\begin{equation}
\mathcal{L} _{} = \lambda _{sdf} \mathcal{L} _{sdf} + \lambda _{normal} \mathcal{L} _{normal} 
\end{equation}

\noindent \textbf{3D Geometric Loss.} Given a sampling point $\mathbf{x}\in \mathcal{X}$, we employ the L2 loss between the predicted $\mathcal{F}_{sdf} (\mathbf{x} )$ and the ground truth $\mathcal{G}_{sdf}(\mathbf{x})$, which are truncated by a threshold $\delta$,

\begin{equation}
\mathcal{L} _{sdf} = \frac{1}{N_S} \sum_{\mathbf{x}\in \mathcal{X}  }^{}\left \| \mathbb{C}(\mathcal{F}_{sdf} (\mathbf{x} ) , \delta ) - \mathbb{C}(\mathcal{G}_{sdf}(\mathbf{x}), \delta ) \right \|_{2}.
\end{equation}

Here $\mathbb{C}(\cdot , \delta ) = max(-\delta, min(\cdot , \delta))$.

\noindent \textbf{Normal Constraint Loss.} Beyond the geometric loss, to make the predicted surface smoother, we deploy Eikonal loss\cite{gropp2020implicit} to encourage the gradients of the sampling points to be close to 1. 

\begin{equation}
\mathcal{L} _{normal} = \frac{1}{N_S}  \sum_{\mathbf{x}\in \mathcal{X}  }^{}\left \| \left \| \nabla _{\mathbf{x} }  \mathcal{F}_{sdf} (\mathbf{x})   \right \|_{2} -1  \right \|_{2}.
\end{equation}



\section{Experiments}

\subsection{Datasets}
We use the WCPA~\cite{dataset2022} dataset as training and testing datasets, which consists of 200 subjects for training and 50 subjects for testing. Each subject contains 15 actions, and each action contains 8 RGB images from different angles (0, 45, 90, 135, 180, 225, 270, 315). Each image is an 1280$\times$720 jpeg file. The ground truth of the canonical pose is a high-resolution 3D mesh with detailed information about clothes, faces, etc. For the training phase, we randomly select 4 RGB images with different views and poses as inputs to learn 3D human models. 

\subsection{Image Preprocessing}

\noindent \textbf{Image Cropping.}
For each image, we first apply VarifocalNet~\cite{zhang2021varifocalnet} to detect the bounding boxes to localize the humans. Next, we crop the input images with the resolution of 512$\times$512 according to the bounding boxes. When the cropped images exceed the bounds of the input images, the cropped images will be padded with zeros.

\noindent \textbf{Mask Generation.}
For each image, we first apply DensePose~\cite{guler2018densepose} to obtain part segmentations. Then we set the parts on the human as the foreground human mask and set the values of the background image pixels as zero. The masked images are served as the input of our network.

The mask is then refined using the MatteFormer \cite{Park_2022_CVPR}, which can generate better boundary details. Following~\cite{wang2021video}, the trimap adopted in~\cite{Park_2022_CVPR} is generated based on the mask using the erosion and dilation operation.

\begin{figure}[t]
\centering
\includegraphics[width=0.98\linewidth]{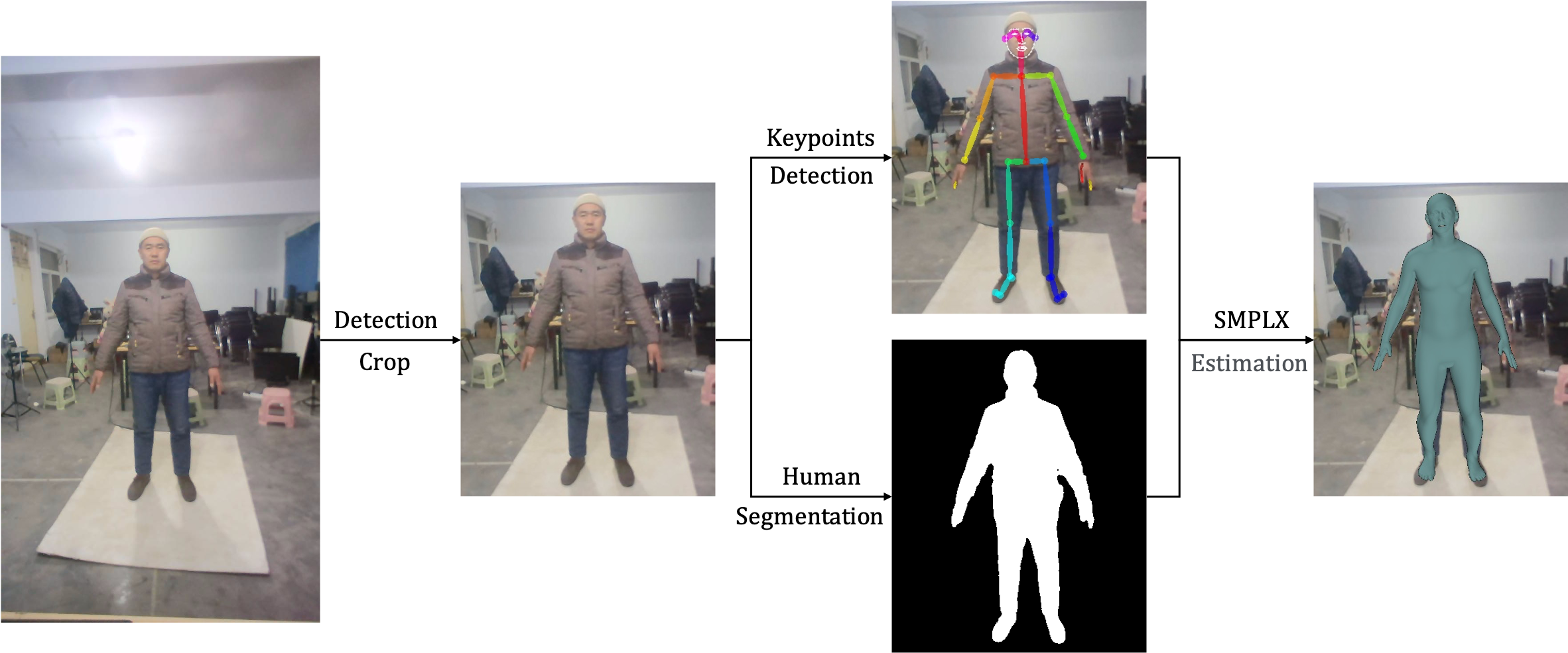}
\vspace{-2mm}
\caption{\textbf{The pipeline for data processing.}}
\label{fig:data_process}
\end{figure}

\subsection{SMPLX Estimation and Optimization}

It is very challenging to get accurate SMPLX from 2D RGB images due to the inherent depth ambiguity. First, we use Openpose\cite{openpose} to detect the 2D keypoints of the person in the image and then use ExPose\cite{expose} to estimate the SMPLX parameters and camera parameters. However, due to extreme illumination conditions and complex backgrounds, the SMPLX obtained in this way is not sufficiently accurate, and we need to refine the SMPLX parameters further. We utilize the 2D keypoints and masks to optimize SMPLX parameters $\theta, \beta$. For the 2D keypoints loss, given the SMPLX 3D joints location $J(\theta, \beta)$, we project them to the 2D image using the weak-perspective camera parameters $s,t$. For the mask loss, we utilize PyTorch3D\cite{ravi2020accelerating} to render the 2D mask given the posed mesh $M(\theta, \beta)$. Then the mask loss is calculated based on the rendered mask and the pseudo ground truth mask $\mathcal{M}_{g t}$. 
\begin{equation}
\theta^{*}, \beta^{*}=\min _{\theta, \beta} \left\|\Phi(J(\theta, \beta), s, t)-\mathcal{J}_{g t}\right\| + \lambda \left\|\Psi(M(\theta, \beta), s, t)-\mathcal{M}_{g t}\right\|
\end{equation}

\begin{figure}[t]
\centering
\includegraphics[width=1.0\linewidth]{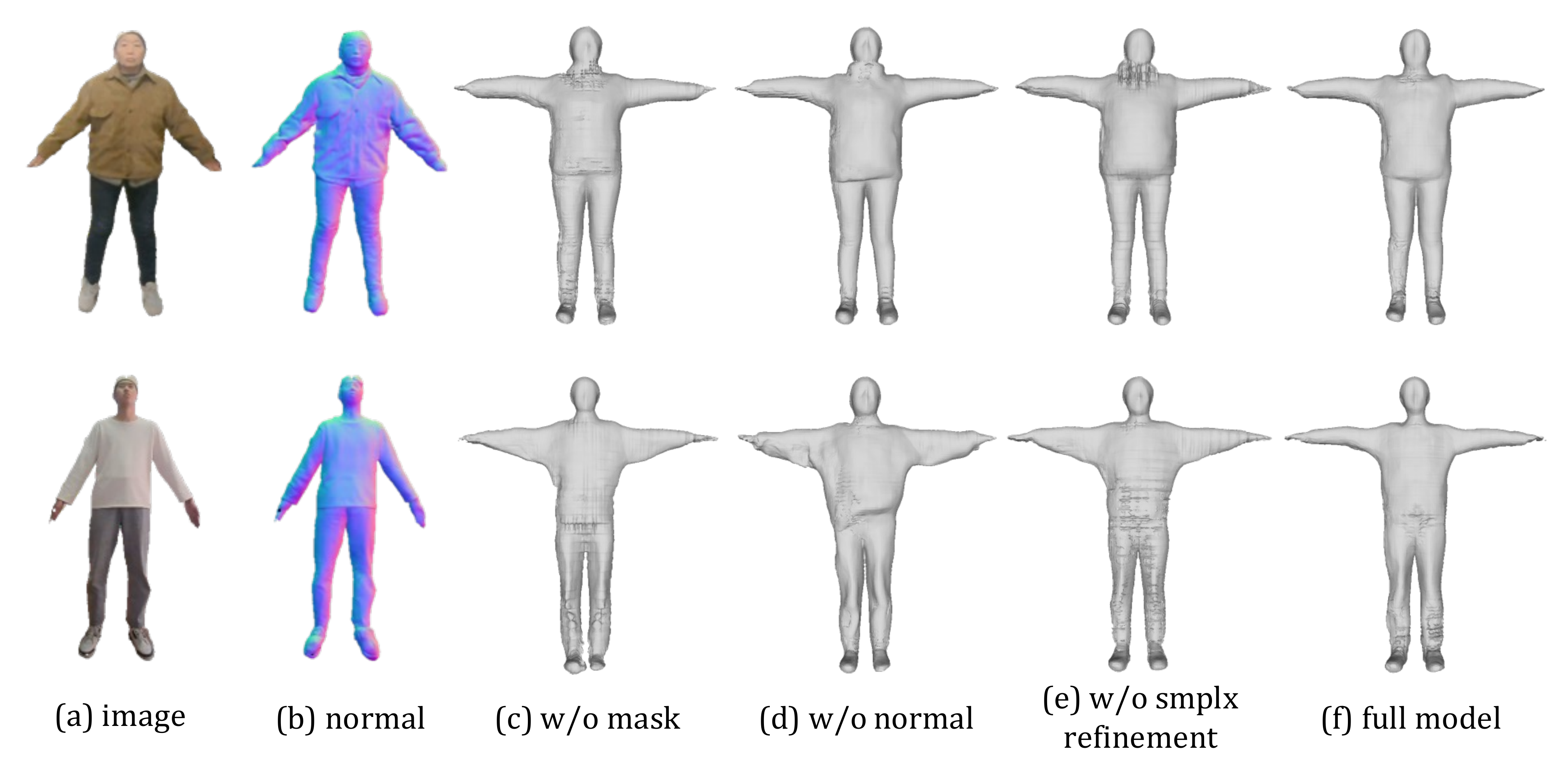}
\vspace{-10mm}
\caption{\textbf{The visualization results on the test dataset.}}
\label{fig:results}
\end{figure}

\subsection{Implementation Details}
During training, we randomly choose $K=4$ images from total of $8*15$ images as input to extract the 2D feature map. Taking into account the memory limitation, we set the latent feature volume $\mathcal{V} \in \mathbb{R}^{224 \times 224 \times 224 \times 64}$, where each latent code $\boldsymbol{l}^{i} \in \mathbb{R}^{64}$. During training, we randomly sample $N_{t} = 10,000$ points around the complete mesh. To stably train our network, we initialize the SDF to approximate a unit sphere~\cite{yariv2020multiview}. We adopt the Adam optimizer~\cite{kingma2014adam} and set the learning rate $lr = 5e-4$, and it spends about 40 hours on 2 Nvidia GeForce RTX 3090 24GB GPUs.
For inference, the surface mesh is extracted by the zero-level set of SDF by running marching cubes on a $256^3$ grid.

\subsection{Results}
Our method achieved very good results in the challenge, demonstrating the superiority of our method for 3D human reconstruction from multi-view and multi-pose images. To further analyze the effectiveness of our method, we performed different ablation experiments. According to the results as shown in the table~\ref{tab:albation_study} and the figure~\ref{fig:results}, removing the background allows the model to better reconstruct the human body. The two-stage approach of predicting the normal vector map as input has a stronger generalization line on unseen data compared to directly using the image as input. What is more important is that the precision of SMPLX has a significant impact on the reconstruction performance, demonstrating the necessity of SMPLX optimization.

\begin{table*}
	\centering
	\scalebox{0.85}{
		\begin{tabular}{c|c|c|c|c}
			\toprule
			 & w/o mask & w/o normal map & w/o SMPLX refinement & full model \\
			\midrule
			Chamfer Distance $\downarrow$ & 1.1277 & 1.0827 & 1.1285 & \textbf{0.9985} \\
			\bottomrule
	\end{tabular}}
	\vspace{0.5em}
	\caption{Quantitative metrics of different strategies on the test dataset.}
	\label{tab:albation_study}
\end{table*}
\vspace{-1.0cm}




\section{Conclusion}

Modeling 3D humans accurately and robustly from challenging multi-views and multi-posed RGB images is a challenging problem, due to the varieties of body poses, viewpoints, light conditions, and other environmental factors. To this end, we have contributed a deep-learning based framework to incorporate the parametric SMPLX model and non-parametric implicit function for reconstructing a 3D human model from multi-images. Our key idea is to overcome these challenges by constructing structured latent codes as the inputs for implicit representation. The latent codes integrate the features for vertices in the canonical pose from different poses or views.
\clearpage
%
%
\bibliographystyle{splncs04}
\bibliography{references}
\end{document}